\documentclass[conference]{IEEEtran}
\IEEEoverridecommandlockouts
\usepackage{cite}

\usepackage{hyperref}
\hypersetup{
    colorlinks=true,
    linkcolor=blue,
    filecolor=black,      
    urlcolor=blue,
}
\usepackage{subcaption}
\usepackage{tabularx,booktabs}
\usepackage{amsmath,amssymb,amsfonts}
\usepackage{xcolor}
\usepackage{algorithmic}
\usepackage{graphicx}
\usepackage{textcomp}
\usepackage{xcolor}
\def\BibTeX{{\rm B\kern-.05em{\sc i\kern-.025em b}\kern-.08em
    T\kern-.1667em\lower.7ex\hbox{E}\kern-.125emX}}
\begin{document}

\title{A Named Entity Based Approach to Model Recipes}
%Potential Titles
% 1. Recipe Entity Recognition : A Named Entity Based Approach to Model Recipes
% 2. 

\author{
\IEEEauthorblockN{1\textsuperscript{st} Nirav Diwan}
\IEEEauthorblockA{\textit{Dept. of Computer Science} \\
\textit{IIIT-Delhi} \\
New Delhi, India \\
nirav17072@iiitd.ac.in}
\and
\IEEEauthorblockN{1\textsuperscript{st} Devansh Batra}
\IEEEauthorblockA{\textit{Dept. of  Information Technology} \\ 
\textit{NSUT}\\
New Delhi, India \\
devanshb.it.17@nsit.net.in}

\and
\IEEEauthorblockN{Ganesh Bagler}
\IEEEauthorblockA{\textit{Center for Computational Biology} \\
\textit{IIIT-Delhi}\\
New Delhi, India \\
bagler@iiitd.ac.in}

}

\maketitle

\begin{table*}
 \caption{\textit{Results of Annotations on the Ingredients Section by the Named Entity Recognition Model}}
\label{table:ingredient-classification}
\begin{tabularx}{\textwidth}{@{}l*{10}{c}c@{}}
\toprule
Ingredient Phrase & Name & State & Quantity & Unit & Temperature & Dry/Fresh & Size  \\ 
\midrule
1 sheet frozen puff pastry ( thawed )   & puff pastry & thawed frozen & 1 & sheet & frozen &  &    \\ 
6 ounces blue cheese,at room temperature & blue cheese &  & 6 & ounces &  &  &  \\
1 tablespoon whole milk ( or half-and-half ) & milk &  & 1 & tablespoon &  &  & \\
2-3 medium tomatoes & tomato &  & 2-3 &  &  &  & medium \\
1/2 teaspoon pepper,freshly ground & pepper & ground & 1/2 & teaspoon &  &  & \\
1/2 teaspoon fresh thyme,minced & thyme & minced & 1/2 & teaspoon &  & fresh & \\
1 teaspoon extra virgin olive oil & extra virgin olive oil &  & 1 & teaspoon &  &  & \\
\bottomrule
\end{tabularx}
\end{table*}

\begin{abstract}
Traditional cooking recipes follow a structure which can be modelled very well if the rules and semantics of the different sections of the recipe text are analyzed and represented accurately. We propose a structure that can accurately represent the recipe as well as a pipeline to infer the best representation of the recipe in this uniform structure. The Ingredients section in a recipe typically lists down the ingredients required and corresponding attributes such as quantity, temperature, and processing state. This can be modelled by defining these attributes and their values. The physical entities which make up a recipe can be broadly classified into utensils, ingredients and their combinations that are related by cooking techniques. The instruction section lists down a series of events in which a cooking technique or process is applied upon these utensils and ingredients. We model these relationships in the form of tuples. Thus, using a combination of these methods we model cooking recipe in the dataset RecipeDB\cite{b1} to show the efficacy of our method. This mined information model can have several applications which include translating recipes between languages, determining similarity between recipes, generation of novel recipes and estimation of the nutritional profile of recipes. 
For the purpose of recognition of ingredient attributes, we train the Named Entity Relationship (NER) models and analyze the inferences with the help of K-Means clustering.
Our model presented with an F1 score of 0.95 across all datasets.
We use a similar NER tagging model for labelling cooking techniques (F1 score = 0.88) and utensils (F1 score = 0.90) within the instructions section. 
% We also identify the temporal sequence of processes in a recipe.
Finally, we determine the temporal sequence of relationships between ingredients, utensils and cooking techniques for modeling the instruction steps.
% Finally, we use Parts of Speech (POS) tagging along with conditional classes to obtain relationships between ingredients, processes and utensils which accurately model the events within the instructions.
\end{abstract}


\section{Introduction}
Instructional language, comprising of step-by-step instructions to be performed in order to complete a task, is a very commonly used structure.
%Instructional language is widely prevalent in the practical world. These are the step-by-step instructions to be performed in order to complete a task. 
Much work has been done to model the instructional prose~\cite{b2}. 
%A lot of work has already been done to model instructional prose well \cite{b3}. 
Cooking recipes have traditionally been considered as a part of such instructional language paradigm, and specialised techniques have been developed to model their representation~\cite{b3}. 
% Cooking recipes have traditionally been associated as a form of this purely instructional language paradigm, and specialised techniques have been developed to model the representation in the same way \cite{b4}. 
However, these techniques require labeled data. Other techniques use complex physical simulators in order to achieve lower level knowledge mining. Lately, unsupervised approaches specific to recipe datasets have caught traction~\cite{b4}; however even these approaches  model recipes as purely instructional in nature. 

\begin{figure}{} %this figure will be at the right
    \centering
    \includegraphics[width=1\columnwidth]{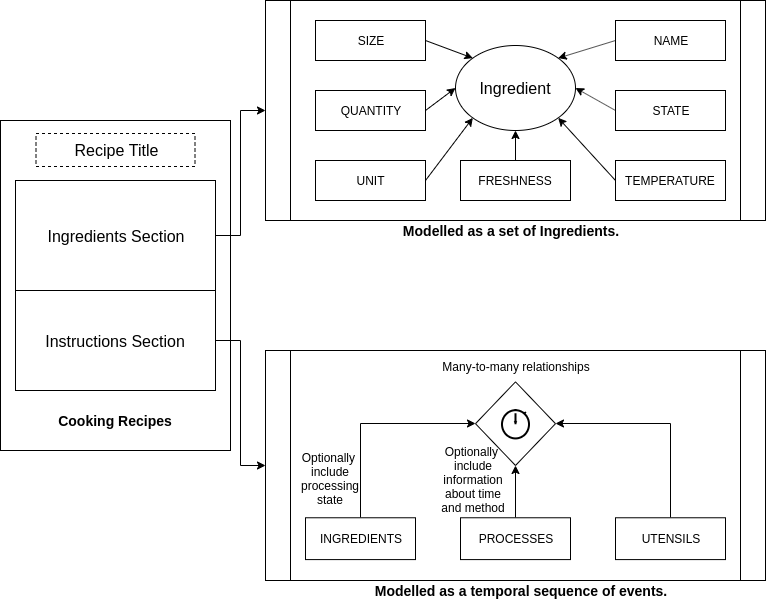}
    \caption{\textit{Diagrammatic representation of  the proposed Recipe Data Structure}}
    \label{fig:recipe_data_structure}
\end{figure}

We challenge this assumption because of the lack of importance given to the ingredients section of the recipe. For advanced analysis of computable semantic insights from a recipe, the ingredients section is of utmost importance and should not be modelled like the rest of the instructional text to prevent data loss. Ingredient information itself can have use cases such as food pairing, flavor prediction, nutritional estimation, cost estimation and cuisine prediction. Therefore, there is a need to accurately represent the ingredient section in a computable format which can be further re-used for other applications similar to instructions.
There are also a lot of differences between the natural language that the instruction sections follow and the ingredients section of the recipe which are devoid of known grammatical structures and rules. Thus, it can be understood why the modeling of both these sections must be done separately and using different techniques in order to get better inferences from the data, unlike the previous studies in this domain.

Fig.~\ref{fig:recipe_data_structure} presents various elements of our model used to represent both the instructions section and ingredients section of cooking recipes. 
A close look on the ingredients section presents it as a list of the ingredients and their corresponding attributes such as quantity, temperature and processing state.
Our aim is to represent the ingredients section of recipes as a structure listing the various attributes. We explain the reasoning behind the attributes selected and our method of computation for the same as well as an analysis of how semantically meaningful these quantities are in the next section. 
Like the traditional approaches, we consider the instructions section to be comprised of a sequence of steps which provide a details of cooking events. We characterize the events as many-to-many relationships between three entities: utensils, processes and cooking techniques. Note that it is not necessary for all the three entities to co-occur. We provide more details on our approach for inferring these relationships in a subsequent section titled `Knowledge Mining from Instruction Section'. 
This information can be used for further applications by interpreting Knowledge Graphs and Thought Graphs from such relationships, themes that have been well explored recently~\cite{b5}.

We acquired the dataset of 118,000 Recipes from RecipeDB\cite{b1}. This database comprises of recipes from primarily two websites AllRecipes.com and FOOD.com (formerly GeniusKitchen.com). 
We constructed our models, conducted subsequent analysis and testing based on this dataset.
All data and code used in this paper are available online on GitHub.~\footnote[1]{https://github.com/cosylabiiit/recipe-knowledge-mining}
\section{Knowledge Mining From Ingredients Section}

\subsection{Challenges in identifying Ingredient and its Attributes}

To extract ingredients and  their corresponding attributes, we primarily investigate the ingredients section of a recipe. 
Ingredients occur in large variety across different recipes. The semantics in writing a recipe vary from region to region. In our database of recipes (118,000) from various regions across the world, we observe there exist primarily the following three challenges in extracting all ingredients and their associated attributes.

\begin{enumerate}
    \item \textbf{Corpus of Attributes}: 
    There is no known corpus which can identify all ingredients or all processing  states in different cooking techniques. 
    Recipes, by nature, are constantly  evolving along with  new ingredients and  new cooking techniques. The model needs to be robust to identify unknown ingredients and  unknown attributes.
    
     \item  \textbf{Identification of Attributes:} Attributes may be homographs. For example, ``clove'' (1 Clove) may refer to the ingredient itself or may refer to the unit of measurement (1 clove of cheese).
    
    \item  \textbf{Variation in Lexical Structure of Ingredient Phrases:} There exist a huge variation in the lexical structure of writing a ingredient phrase. The structure may be as simple as ``3⁄4 cup sugar'' OR ``1 garlic clove, crushed'' to being as complex as ``1 (8 ounce) package cream cheese, softened'' or ``1-2 fresh chili pepper very finely chopped.''
\end{enumerate}

\subsection{Named Entity Recognition Model}
Since the recipes evolve in terms of their ingredients and the corresponding attributes, understanding the structure of the ingredient phrases is critical. 

Named Entity Recognition (NER) models are generally trained to  automate the task of annotating words with tags. Hence, we propose training a NER model with the help of a manually annotated subset of recipes.
The named entities we propose are-- Name, Unit, Quantity, Processing State, Size, Dryness/Freshness state, and Temperature. These entities have been proposed after manually going through descriptions in the Standard Legacy Database by USDA, one of the recommended reference databases for nutritional information. We describe  each of the tags in Table~\ref{table:NER_Tags}.

We train the Stanford NER Tagger\cite{b6} and utilize it to annotate the ingredient sections of our database of 118,000 recipes. For more details of the training process and conducted refer II.F. Consider the recipe Tomato and Blue Cheese Tart~\footnote[2]{https://www.food.com/recipe/heirloom-tomato-blue-cheese-tart-325721}. Table~\ref{table:ingredient-classification} shows the results of applying the trained NER model for seven ingredient phrases. 

\subsection{Pre-Processing}
Before passing the ingredient phrase to the NER model, we pre-process the ingredient phrases by eliminating stop words, lemmatizing all words using WordNet Lemmatizer, and converting the text to lower case. This pre-processing has advantages like correctly classifying ingredient terms which differ in plurality, capitalization, presence of hyphens as identical entities. For instance, both ``tomatoes'' and ``Tomato'' are output as ``tomato'', thus making the generated data much more useful for different purposes. As part of our implementation of the framework, we used the NLTK library for the purpose of pre-processing.

\subsection{Parts of Speech (POS) Tagging to 
represent Vectors}
As a preliminary experiment, we noticed that using a small set of annotated examples for training was not successful in accurately identifying the entities in the entirety of our diverse dataset.
We attribute this to the presence of a number of lexical structures of the ingredient phrases which vary widely across cuisines and original data sources, mentioned earlier as a challenge. 

To efficiently tackle this problem, we devise a novel technique for finding the Parts of Speech (POS) Tags for these phrases. These Parts of Speech Tags assign tags to each token, such as noun, verb, adjective, etc. The mainstream sophisticated taggers available provide additional fine-grained POS tags such as `noun-plural'. 

In our experiment, we use the Stanford POS Twitter model~\cite{b7} in order to find the POS tags for all ingredient phrases in the RecipeDB database. The intuition behind using this tagger is that ingredient phrases are not grammatically complete sentences, but resemble tweets in their lexical structures. We represented the output for every unique ingredient phrase as vectors ($1 \times 36$) on the basis of the frequency of each POS tag present.
The corresponding vectors for ingredient phrases with similar lexical structures will be close to each other in terms of euclidean distance. This intuition is backed by dimensional interrelationships from models like Word2Vec which is formed on a similar basis. 

\begin{figure*}{} %this figure will be at the right
    \centering
    \includegraphics[width=2\columnwidth]{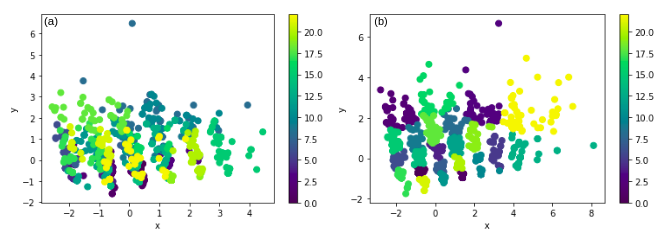}
    
    \caption {
    \textit{
     (a) Visualization of the Clusters formed by applying k - means Clustering on the vectors formed by determining the frequency of POS Tags in the ingredient phrase. \textbf{After clustering the vectors}, the visualization has been generated by applying Principal Component Analysis on the vectors for converting 36 - dimensional vectors into 2 dimensions. The clusters are separable at higher dimensions (b) Visualization of the clusters formed by applying k - means Clustering on the vectors formed by determining the frequency of POS Tags\cite{b11} in the ingredient phrase. \textbf{Before clustering the vectors}, we apply Principal Component Analysis on the vectors for converting 36 - dimensional vectors into 2 dimensions. We then proceed to cluster the vectors. 
     The visualizations display 50 unique ingredient phrases from each of the 23 distinct clusters
        }
    \label{fig:clusters}
     }

\end{figure*}

% \begin{figure}{} %this figure will be at the right
%     \centering
%     \includegraphics[width=1\columnwidth]{PCA_before_clustering.png}
% \end{figure}

\subsection{Analysis of vectors by clustering}
We further create clusters within this vector space. The constituent ingredient phrases within a cluster are likely to have similar lexical structures. We can now pick a small subset of sentences such that it has a sufficient number of representatives from each of the clusters. These representative phrases form a good training set for the NER tagger, provided that the number of chosen phrases are sufficient and hyperparameters of the clustering algorithm have been optimised.

For the AllRecipes (16,000 Recipes) corpus (1.5 million of 11.5 million ingredient phrases), we grouped ingredient phrases into clusters. From each cluster, 1\% unique ingredient phrases were picked which were used to form a training set of 1,500 ingredient phrases. Similarly, 0.33\% of unique ingredient phrases (specifically excluding the ingredient phrases in the training set) were then picked which were used then used to form a testing set of 500 ingredient phrases.
The same methodology was applied for the Food.com (102,000 recipes) corpus which had 10 million of the 11.5 million ingredient phrases. However, since the number of total ingredient phrases were much larger than the AllRecipes corpus (about 10 times as large) we proceeded to select 0.5\% from each cluster to form the training set of 5000 ingredient phrases and to select 0.165\% from each cluster to form the testing set of 1,500 ingredient phrases.
Combining the above two corpora, we obtained a training set of 6,612 ingredient phrases and a testing set of 2188 ingredient phrases.

K-Means Clustering Algorithm was utilized to cluster the ingredient vectors. The clusters are formed on the basis of the frequency of the tags through a Bag-of-Words Approach.
The decision of selecting the number of clusters formed was based on two factors: inertia of the clusters formed (Elbow Criterion Method\cite{b8}) and interpretation of the clusters. For example, the phrases ``3 teaspoons olive oil'' and ``2 tablespoons all-purpose flour'', ideally, should belong to the same cluster since they correspond to form a similar sentence structure as they both contain one cardinal number, two nouns and one plural-noun. Based on these factors, \textbf{23 distinct clusters (Fig~\ref{fig:clusters})} were identified. The ingredient phrases were then manually tagged as described in Table~\ref{table:NER_Tags}.

%-------------------TABLE 2-------------------------------
\begin{table}[htbp]
\caption{Named Entity Recognition Tags}
\begin{center}
\begin{tabular}{|c|c|c|}
\hline
\textbf{Tag}&{\textbf{Significance}}  & \textbf{Example} \\
\cline{2-3} 
% \textbf{Tag} & \textbf{\textit{Meaning}}&  \\
% \textbf{\textit{Subhead}}& \textbf{\textit{Subhead}} \\

\hline
& & \\
NAME &  Name of Ingredient & salt, pepper\\
STATE  & Processing State of Ingredient. & ground, thawed \\
UNIT & Measuring unit(s) & gram, cup\\
QUANTITY & Quantity associated with the unit(s). &  1, 1$\frac{1}{2}$, 2-4\\
SIZE &  Portion sizes mentioned & small, large\\
TEMP & Temperature applied prior to cooking & hot, frozen\\
DRY/FRESH & Fresh otherwise as mentioned. & dry, fresh\\ 
\hline

% \multicolumn{4}{l}{$^{\mathrm{a}}$Sample of a Table footnote.}
\end{tabular}
\label{table:NER_Tags}
\end{center}
\end{table}

%-------------------TABLE 3-------------------------------
\begin{table}[htbp]
\caption{Training and Testing Dataset Sizes For NER on Ingredients Section}
\begin{center}
\begin{tabular}{|c|c|c|c|}
\hline

\textbf{Datasets} &{\textbf{AllRecipes}}  & \textbf{FOOD.com}  & \textbf{BOTH}\\
\cline{2-4} 
% \textbf{Tag} & \textbf{\textit{Meaning}}&  \\
% \textbf{\textit{Subhead}}& \textbf{\textit{Subhead}} \\
\hline
& & & \\
Training Set Size & 1470 & 5142  & 6612 \\
Testing Set Size  & 483  & 1705  & 2188 \\
\hline
\end{tabular}
\label{table:dataset-description}
\end{center}
\end{table}

\subsection{Evaluation}
We trained the Stanford NER Model~\cite{b6} based on our three datasets divided into training and testing sets as described in Table~\ref{table:dataset-description}.  We tested the three models on the subsequent testing sets across each datasets as highlighted in Table~\ref{table:NER-Evalaution}. The  models were validated by 5-fold cross validation.
The NER model trained on AllRecipes Dataset gives an F1 Score of 0.9682 on the AllRecipes.com Test Set. Similarly, the NER Model trained of FOOD.com Dataset gives an F1 Score of 0.9519 on the FOOD.com Test Set. However, the individual models do not consistently give the same F1 scores across all datasets. We attribute this to the variation of the recipes between the two datasets. The model trained on the composite dataset of AllRecipes and FOOD.com delivers the best performance with an F1 scores of over 0.95 across the three datasets. 
The model is significant as it allows to derive ingredient and its corresponding attributes given an ingredient phrase from the ingredients section.

From our database of 118,000 recipes, 
Through our model we extracted 
 \textbf{20,280} unique ingredient names. It is important to note that there are a number of aliases of same ingredients included in this final count. For example, okhra and ladyfinger are counted as two different ingredient names although they represent the same ingredient.

%-------------------TABLE 4-------------------------------
\begin{table}[htbp]
    \centering
    \caption{Evaluation of NER Model for Ingredients Section}
    \begin{tabular}{|c|c|c|c|}
        \hline
        \textbf{Testing} &     \multicolumn{3}{c|}{\textbf{Training Set Model}}\\
        \cline{2-4}
        \textbf{Set} & \textit{\textbf{AllRecipes}} & \textit{\textbf{FOOD.com}} & \textit{\textbf{BOTH}} \\ 
        \hline
        AllRecipes  & \textbf{0.9682}  & 0.9317  & \textbf{0.9709} \\
\hline
FOOD.com  & 0.8672  & \textbf{0.9519}  & \textbf{0.9498} \\
\hline
BOTH  & 0.8972  & 0.9472  & \textbf{0.9611} \\
\hline

\end{tabular}
\label{table:NER-Evalaution}

\end{table}

% \section{Instruction Data Extraction}
\section{Knowledge Mining From Instructions Section}

% \subsection{Knowledge Mining From Instructions Section}

In order to help machine understanding of structured sequential instructions data, a lot of approaches have been applied historically~\cite{b9}. However, unsupervised learning in this domain has only caught up recently~\cite{b10}. Our goal is to similarly learn the event chain within a recipe in an unsupervised manner because of the complex and extremely time-consuming task of labelling training data.

We define recipe instructions as being a narrative chain — a sequence of events which is constituted of multiple protagonists interacting together in predefined relationships. The protagonists in a recipe are the utensils and ingredients. We model the relationships using Cooking Techniques/Processes (the act of processing ingredients and utensils such as preheating an oven, or boiling a potato). Thus the task is to identify these temporal relations. For this purpose we follow a two step strategy.

\begin{figure*}{}
\centering\includegraphics[width=2\columnwidth]{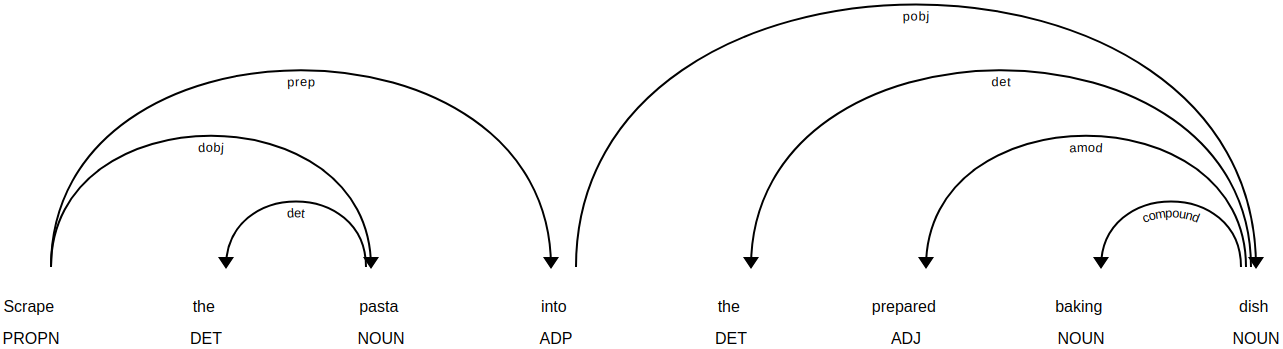}
\caption {
    \textit{
     A dependency parsed structure of a typical instruction using the Spacy library
        }
    \label{fig:spacy-dependency-tree}
     }
\end{figure*}
\subsection{Entity Recognition within Instruction Set}
Firstly, we train another NER model which identifies the Processes, Utensils and Ingredients within the instruction set. The training process was very similar to the one followed for the ingredients section. The recipes with the longest instructions section from 40 different cuisines were extracted to further annotate 268 processes/techniques, ingredients, and 69 utensils. The Stanford NER tagger~\cite{b6} was trained upon this corpus. The model thus obtained was then used for tagging the cooking recipes within RecipeDB.

We used this pre-trained NER model to label entities within the instruction sets of recipes in RecipeDB after pre-processing them in a manner similar to the one implemented for ingredients section. We then used the threshold frequencies (47 and 10 respectively), to create a dictionary of Cooking Techniques and Utensils which was used to filter the data from the NER model, removing most of the inconsistencies. 

\begin{figure*}{} %this figure will be at the right
    \centering
    \includegraphics[width=2\columnwidth]{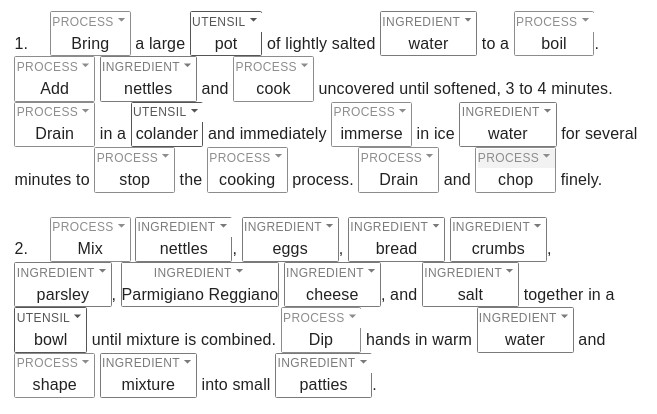}
    
    \caption{
        \textit{
     The inference from the NER model for the instruction section of a recipe
        }
        \label{fig:results_ner}
    }
\end{figure*}

Fig.~\ref{fig:results_ner} shows an example of entity tags inferred using this approach. The performance of the model is explained by Table~\ref{table:ner-evaluation-instructions}.

\begin{table}[htbp]
\caption{Evaluation of NER model for Instructions Section}

\begin{center}

\begin{tabular}{|c|c|c|c|}
\hline

\textbf{F1 } &{\textbf{Precision}}  & \textbf{Recall}  & \textbf{F1 Score}\\
\cline{2-4} 
% \textbf{Tag} & \textbf{\textit{Meaning}}&  \\
% \textbf{\textit{Subhead}}& \textbf{\textit{Subhead}} \\
\hline
& & & \\

\textbf{Processes} & 0.92  & 0.85  & 0.88  \\
\textbf{Utensils}  & 0.94 & 0.86 & 0.90 \\
\hline
\end{tabular}
\label{table:ner-evaluation-instructions}
\end{center}
\textit{
Performance Analysis of Named Entity Recognition Model on Instructions Section for the extraction of Processes and Utensils
}
\end{table}

\subsection{Relationship Extraction}
In order to extract the events from instructions section of cooking recipes, we study the grammatical structure of these recipes to form tuples which can model the section well. Many approaches have already been proposed for dependency parsing and subject-verb extraction for relation extraction~\cite{b6}. We have described below our approach in this regard.
The tuples to be inferred include the entities (both Utensils and Ingredients) and the relationships between these entities. The relationships are defined by the Cooking Techniques being applied on the ingredients and utensils. The relationships are many-to-many for a particular Cooking Technique at an instance. This can be better understood with an example. An instruction may state that potatoes are boiled. It may also state that potatoes are fried with olive oil in a pan. In this case, the relation extends to both the ingredients - olive oil and potatoes as well as the utensil - pan.

% Out of 2086 sentences (extracted from 120 longest recipes from across all cuisines) that were manually annotated, 1586 were randomly chosen for training and the rest (500) were used for testing. The models presented with F1 score of 0.88 (Precision: 0.92 and Recall: 0.85) and 0.90 (Precision: 0.94 and Recall: 0.86), respectively for processes and utensils, respectively. 

\begin{figure}{} 
    \centering
    \includegraphics[width=1\columnwidth]{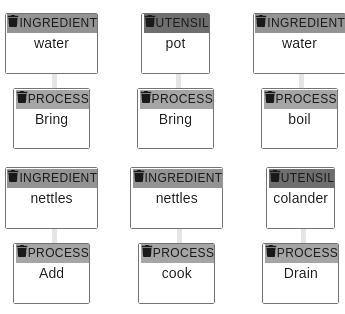}
    \caption{
    \textit{
    The inference from relationship tagging for part of the first instruction in Fig.~\ref{fig:results_ner}. Notice that both Bring + Water and Bring + Pot relations are modeled by the same process - Bring. Multiple such one-one relations are combined together if they are found to be the dependency of the same process to form compound many-to-many relations
    \label{fig:results-relation-extraction}
    }}
\end{figure}

Fig~\ref{fig:spacy-dependency-tree} shows the dependency parsing tree obtained using the Spacy library~\cite{b12}. Our aim is to find out the relationships using grammatical rules. It is easy to follow from intuition that the Cooking Techniques are generally tagged as verb in any instruction. Thus, for all the verbs classified as processes (having filtered after pre-processing and comparing with the terms in the dictionary mentioned before), we attempt to find the associated subjects and objects within the dependency tree. We also obtain prepositional objects. These are likely to be the processes and utensils associated with the relationship. In order to satisfy the many-to-many entity criteria, we also filter this list of relationships using the NER inferred Ingredients and Utensils to obtain the final relationships. Fig. \ref{fig:results-relation-extraction} shows an example of relationships obtained using this approach.

% Out of 2086 sentences (extracted from 120 longest recipes from across all cuisines) that were manually annotated, 1586 were randomly chosen for training and the rest (500) were used for testing. The models presented with F1 score of 0.88 (Precision: 0.92 and Recall: 0.85) and 0.90 (Precision: 0.94 and Recall: 0.86), respectively for processes and utensils, respectively. 

\section{Applications and Future Scope}

The proposed data structure captures a significant part of the cooking recipe:
Ingredients and their associated characteristics; Cooking processes utilised in the recipe; Temporal sequence of cooking processes; Relationship between cooking processes and ingredients; Utensils utilised in the recipe; and Temporal relationships between utensils, ingredients and processes.

Such a structured information is of key value for researchers. A similar model which does not take into account the information from Ingredients Section has been useful in tasks like Artificial Intelligence based Recipe Text Generation~\cite{b4}. Our model could help to extract information out of the Ingredients Section synchronised with the Instructions Section. We have implemented our model for estimation of nutritional from cooking recipes~\cite{b13}, as well as for finding similar recipes in RecipeDB~\cite{b1}.

\section{Conclusions}

We present a novel method to extract ingredients and its corresponding attributes in a structured format via the utilization of clustering and Named Entity Recognition. We also present a labelled dataset of 8800 Ingredient phrases, divided into training and testing sets~\ref{table:dataset-description}, manually tagged into 7 categories as described in Table~\ref{table:NER_Tags}. Our Best Model gives a F1 Score of  greater than  or equal  to 0.95 across all datasets.

For data extraction from the instruction section, we applied the aforementioned pipeline (from Section III) on 40,000 randomly chosen recipes with 174,932 instruction steps from RecipeDB~\cite{b1}. After filtering out the tuples to only include the relationships defined by processes predicted by the NER tagger, we observed that on average, an instruction yielded 6.164 relations with a standard deviation of 5.70 which shows the large variation in the number of entities a cooking technique may be applied on. The large standard deviation re-affirms the need to model relationships in a many-to-many fashion. Else, a lot of information about an event may have been lost if one-to-many, one-to-one, or many-to-one relations were used for the modeling since they would not have been able to accommodate the large variation in number of ingredients and utensils at the same time.

In summary, we propose the Named Entity based approach to represent recipes. This approach has several applications which include translating recipes between languages, determining similarity between recipes, generation of novel recipes and estimation of nutritional profile of recipes. The proposed model was utilized in RecipeDB~\cite{b1} to determine similarity between recipes and for nutritional estimation of recipes.

% \begin{figure*}{t}
% \centering\includegraphics[width=2\columnwidth]{tree.png}
% \end{figure*}

% \begin{figure}{} %this figure will be at the right
%     \centering
%     \includegraphics[width=1\columnwidth]{Recipe-DS.png}
% \end{figure}

\section*{Acknowledgment}
N.D. and D.B. contributed equally to the work done towards this manuscript. N.D. and D.B. thank IIIT-Delhi and Complex Systems Laboratory for the Summer Research Internship. G.B. thanks IIIT-Delhi for the computational facilities. The authors thank Rudraksh Tuwani for suggesting the use of NER as part of computational protocol.
% \section*{References}

% Please number citations consecutively within brackets \cite{b1}. The 
% sentence punctuation follows the bracket \cite{b2}. Refer simply to the reference 
% number, as in \cite{b3}---do not use ``Ref. \cite{b3}'' or ``reference \cite{b3}'' except at 
% the beginning of a sentence: ``Reference \cite{b3} was the first $\ldots$''

% Number footnotes separately in superscripts. Place the actual footnote at 
% the bottom of the column in which it was cited. Do not put footnotes in the 
% abstract or reference list. Use letters for table footnotes.

% Unless there are six authors or more give all authors' names; do not use 
% ``et al.''. Papers that have not been published, even if they have been 
% submitted for publication, should be cited as ``unpublished'' \cite{b4}. Papers 
% that have been accepted for publication should be cited as ``in press'' \cite{b5}. 
% Capitalize only the first word in a paper title, except for proper nouns and 
% element symbols.

% For papers published in translation journals, please give the English 
% citation first, followed by the original foreign-language citation \cite{b6}.

\end{document}